\pdfoutput=1

\documentclass[11pt]{article}

\usepackage[]{acl}

\usepackage{times}
\usepackage{latexsym}

\usepackage[T1]{fontenc}
\usepackage{ulem}
\usepackage{graphicx}
\usepackage[utf8]{inputenc}

\usepackage{microtype}

\usepackage{inconsolata}

\usepackage{booktabs}
\usepackage{makecell}
\usepackage{multirow}
\usepackage{soul, color, xcolor}
\usepackage{tikz}
\usepackage[edges]{forest}
\usepackage{ulem}

\definecolor{hidden-draw}{RGB}{20,68,106}
\definecolor{hidden-pink}{RGB}{255,245,247}
\definecolor{light-yellow}{RGB}{248, 253, 207}
\definecolor{light-green}{RGB}{226, 246, 202}
\definecolor{light-skyblue}{RGB}{205, 245, 253}
\definecolor{light-blue}{RGB}{120, 193, 243}

\title{A Survey of Large Language Models Attribution}

\author{
    \begin{tabular}{c}
    Dongfang Li$^1$ \quad Zetian Sun$^1$\quad Xinshuo Hu$^1$ \quad Zhenyu Liu$^1$ \quad Ziyang Chen$^2$ \vspace{.5mm} \\
    Baotian Hu$^1$\quad Aiguo Wu$^1$\quad Min Zhang$^1$
    \end{tabular}
    \\ \vspace{.5mm}
    \begin{tabular}{c}
    $^1$ Harbin Institute of Technology (Shenzhen), Shenzhen, China \\
    $^2$ Laboratory for Big Data and Decision, National University of Defense Technology, China\\
    \end{tabular}
    \\ \vspace{.5mm}
    \texttt{\{lidongfang,hubaotian,zhangmin2021\}@hit.edu.cn}
}

\begin{document}
\maketitle
\begin{abstract}
Open-domain generative systems have gained significant attention in the field of conversational AI (e.g., generative search engines). This paper presents a comprehensive review of the attribution mechanisms employed by these systems, particularly large language models. 
Though attribution or citation improve the factuality and verifiability, issues like ambiguous knowledge reservoirs, inherent biases, and the drawbacks of excessive attribution can hinder the effectiveness of these systems. The aim of this survey is to provide valuable insights for researchers, aiding in the refinement of attribution methodologies to enhance the reliability and veracity of responses generated by open-domain generative systems. We believe that this field is still in its early stages; hence, we maintain a repository to keep track of ongoing studies at \url{https://github.com/HITsz-TMG/awesome-llm-attributions}.
\end{abstract}

\section{Introduction}
Since the advent of open-domain generative systems driven by large language models (LLMs)~\cite{Anil2023PaLM2T,chatgpt,OpenAI2023GPT4TR}, addressing the coherent generation of potentially inaccurate or fabricated content has been a persistent challenge~\cite{Rawte2023ASO,Ye2023CognitiveMA,Zhang2023SirensSI}. Such issues are often referred to by the community as ``hallucination'' problems, wherein the generated content presents distorted or invented facts which lacks credible sourcing~\cite{DBLP:conf/acl/PeskoffS23}. This becomes particularly obvious in information-seeking and knowledge question-answering scenarios, where users rely on large language models for expert knowledge~\cite{Malaviya2023ExpertQAEQ}. 
\begin{figure}[t!]
\centering
\includegraphics[width=1.0\linewidth]{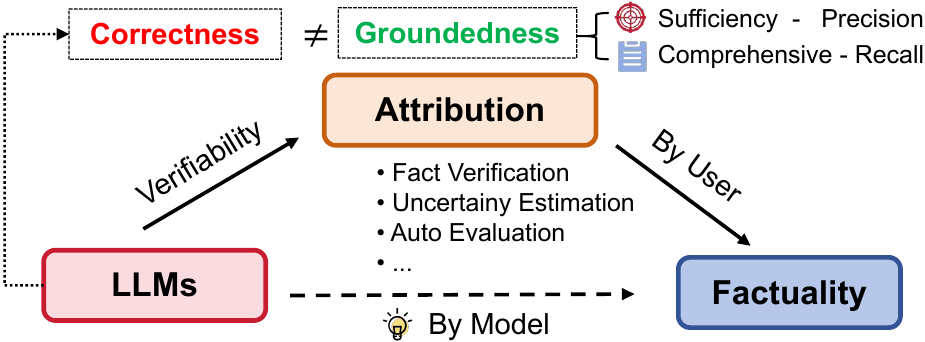}
  \caption{By providing attribution, both developers and users can view the possible source of an answer and evaluate factuality and reliability to form their own assessment. Attribution as a more realistic way to reduce hallucinations bypasses the task of directly determining the ``truthfulness'' of statements, a feat difficult to achieve except for the most basic queries. }
    
   \label{figmotivation}
\end{figure}

\tikzstyle{my-box}=[
    rectangle,
    draw=light-blue,
    rounded corners,
    text opacity=1,
    minimum height=1.5em,
    minimum width=5em,
    inner sep=2pt,
    align=center,
    fill opacity=.5,
    line width=0.8pt,
]
\tikzstyle{leaf}=[my-box, minimum height=1.5em,
    fill=hidden-pink!80, text=black, align=left,font=\normalsize,
    inner xsep=2pt,
    inner ysep=4pt,
    line width=0.8pt,
]
\begin{figure*}[t!]
    \centering
    \resizebox{\textwidth}{!}{
        \begin{forest}
            forked edges,
            for tree={
                grow=east,
                reversed=true,
                anchor=base west,
                parent anchor=east,
                child anchor=west,
                base=left,
                font=\large,
                rectangle,
                draw=light-blue,
                rounded corners,
                align=left,
                minimum width=4em,
                edge+={darkgray, line width=1pt},
                s sep=3pt,
                inner xsep=2pt,
                inner ysep=3pt,
                line width=0.8pt,
                ver/.style={rotate=90, child anchor=north, parent anchor=south, anchor=center},
            },
            where level=1{text width=7em,font=\normalsize,}{},
            where level=2{text width=8.5em,font=\normalsize,}{},
            where level=3{text width=8em,font=\normalsize,}{},
            where level=4{text width=5em,font=\normalsize,}{},
            [
                Large Language Models Attribution, ver, fill=light-yellow
                [
                    Sources (\S \ref{sec:Sources_of_Attribution}), fill=light-green
                    [
                        Pre-training \\Data  (\S \ref{sec:Pre-training_Data}), fill=light-skyblue
                        [
                            ORCA~\cite{DBLP:journals/corr/abs-2205-12600}{, }
                            ORCA-ICL~\cite{DBLP:conf/acl/HanSMTCW23}{, }\\
                            ROOTS~\cite{DBLP:conf/acl/PiktusAVLDLJR23}{, }
                            QUIP-Score~\cite{DBLP:journals/corr/abs-2305-11859}
                            , leaf, text width=29em
                        ]
                    ]
                    [
                        Out-of-model \\Knowledge  (\S \ref{sec:Out-of-model_Knowledge}), fill=light-skyblue
                        [
                            RR~\cite{DBLP:journals/corr/abs-2301-00303}{,}
                            Retrieval Augmentation~\cite{DBLP:conf/emnlp/0001PCKW21}{, }\\
                            Retro~\cite{DBLP:conf/icml/BorgeaudMHCRM0L22}{, }
                            FiD~\cite{DBLP:conf/eacl/IzacardG21}
                            , leaf, text width=29em
                        ]
                    ]
                ]
                [
                    Datasets (\S \ref{sec:Datasets_for_Attribution}), fill=light-green
                    [
                        WebGPT~\cite{nakano2021webgpt}{, }
                        WebBrain~\cite{DBLP:journals/corr/abs-2304-04358}{, }
                        WebCPM~\cite{qin-etal-2023-webcpm}{, }\\
                        CiteBench~\cite{https://doi.org/10.48550/arxiv.2212.09577}{, }
                        ALCE~\cite{gao2023enabling}{, }
                        ExpertQA~\cite{Malaviya2023ExpertQAEQ}{,}\\
                        HAGRID~\cite{hagrid}{, }
                        SEMQA~\cite{DBLP:journals/corr/abs-2311-04886}{, }
                        BioKaLMA~\cite{li2023verifiable}
                        , leaf, text width=43em
                    ]
                ]
                [
                    Approaches (\S \ref{sec:Approaches_to_Attribution}), fill=light-green
                    [
                        Direct Generated \\Attribution  (\S \ref{sec:Direct_Generated_Attribution}), fill=light-skyblue
                        [
                            RECITE~\cite{DBLP:conf/iclr/Sun0TYZ23}{, }
                            Blueprint Model~\cite{anonymous2023learning}{, }\\
                            according-to Prompting~\cite{weller2023according}{, }
                            IFL~\cite{lee2023towards}{,}\\
                            Attributing Prompting~\cite{zuccon2023chatgpt}{, }
                            1-PAGER~\cite{DBLP:journals/corr/abs-2310-16568}{, }\\
                            Credible Without Credit~\cite{DBLP:conf/acl/PeskoffS23}
                            , leaf, text width=32em
                        ]
                    ]
                    [
                        Post-retrieval \\Answering  (\S \ref{sec:Post-retrieval_Answering}), fill=light-skyblue
                        [
                            SearChain~\cite{DBLP:journals/corr/abs-2304-14732}{, }
                            MixAlign~\cite{DBLP:journals/corr/abs-2305-13669}{, }\\
                            SmartBook~\cite{DBLP:journals/corr/abs-2303-14337}{, }
                            Self-RAG~\cite{DBLP:journals/corr/abs-2310-11511}{, }\\
                            LLatrieval~\cite{li2023llatrieval}{, }
                            AGREE\cite{ye2023effective}
                            , leaf, text width=29em
                        ]
                    ]
                    [
                        Post-Generation \\Attribution  (\S \ref{sec:Post-Generation_Attribution}), fill=light-skyblue
                        [
                            RARR~\cite{DBLP:conf/acl/GaoDPCCFZLLJG23}{, }
                            SAPLMA~\cite{DBLP:journals/corr/abs-2304-13734}{, }\\
                            IQ DQ~\cite{DBLP:journals/corr/abs-2305-18248}{,}
                            CCVER~\cite{DBLP:journals/corr/abs-2305-11859}{,}\\
                            RSEQGA~\cite{DBLP:journals/corr/abs-2309-11392}
                            , leaf, text width=29em
                        ]
                    ]
                    [
                        Attribution Systems  \\(\S \ref{sec:Attribution_Systems}), fill=light-skyblue
                        [
                            LaMDA~\cite{Thoppilan2022LaMDALM}{, }
                            WebGPT~\cite{nakano2021webgpt}{, }\\
                            GopherCite~\cite{Menick2022TeachingLM}{, }
                            Sparrow~\cite{DBLP:journals/corr/abs-2209-14375}{, }\\
                            WebCPM~\cite{qin-etal-2023-webcpm}
                            , leaf, text width=29em
                        ]
                    ]
                ]
                [
                    Evaluation (\S \ref{sec:Attribution_Evaluation}), fill=light-green
                    [
                        Side~\cite{DBLP:journals/corr/abs-2207-06220}{, }
                        Auto-AIS~\cite{DBLP:conf/acl/GaoDPCCFZLLJG23}{, }
                        \citet{DBLP:journals/corr/abs-2212-08037}{, }
                        \citet{Liu2023EvaluatingVI}{,}\\
                        WICE~\cite{DBLP:journals/corr/abs-2303-01432}{, }
                        XOR-AttriQA~\cite{DBLP:journals/corr/abs-2305-14332}{, }
                        FActScore~\cite{DBLP:journals/corr/abs-2305-14251}{, }\\
                        AttrScore~\cite{DBLP:journals/corr/abs-2305-06311}{, }
                        FacTool~\cite{DBLP:journals/corr/abs-2307-13528}{, }
                        \citet{DBLP:journals/corr/abs-2309-05217}{, }
                        KaLMA~\cite{li2023verifiable}
                        , leaf, text width=44em
                    ]
                ]
            ]
        \end{forest}
    }
    \caption{Taxonomy of Large Language Models Attribution.}
    
    \label{taxo_of_icl}
\end{figure*}
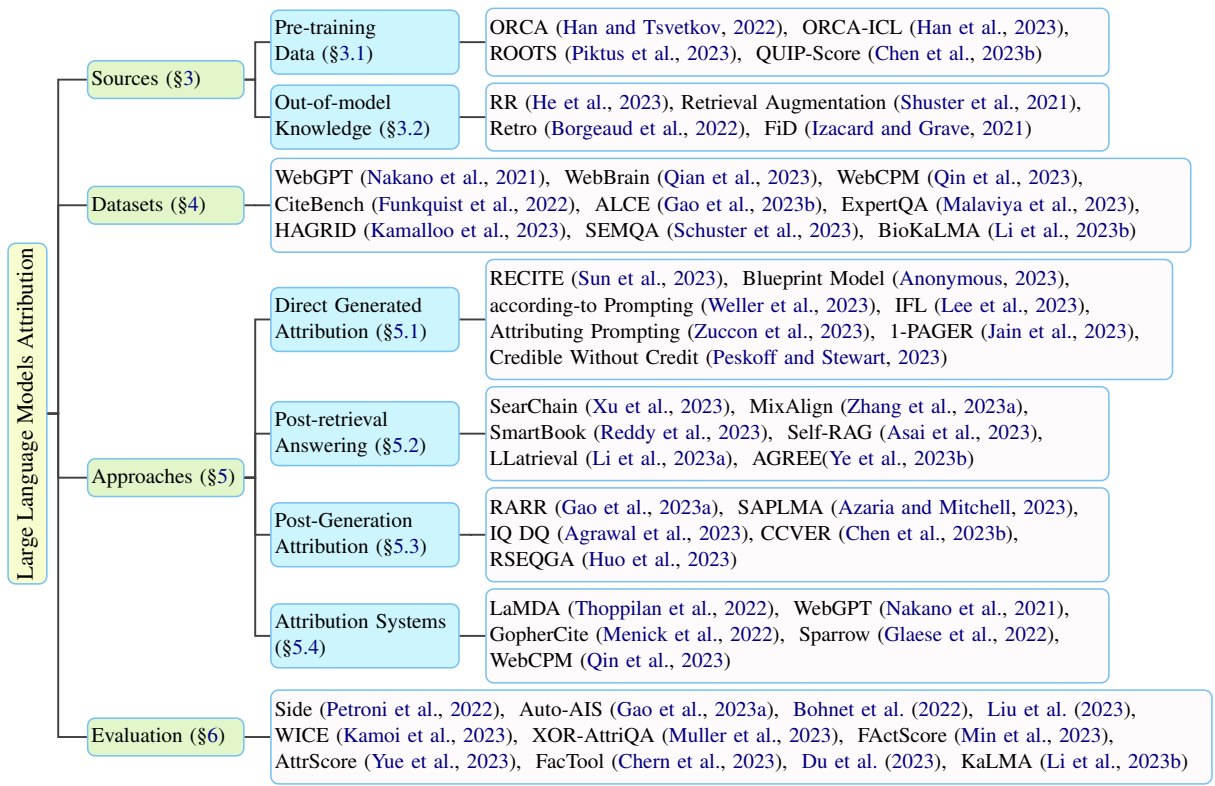

\begin{figure*}[t!]
\centering
\includegraphics[width=\linewidth]{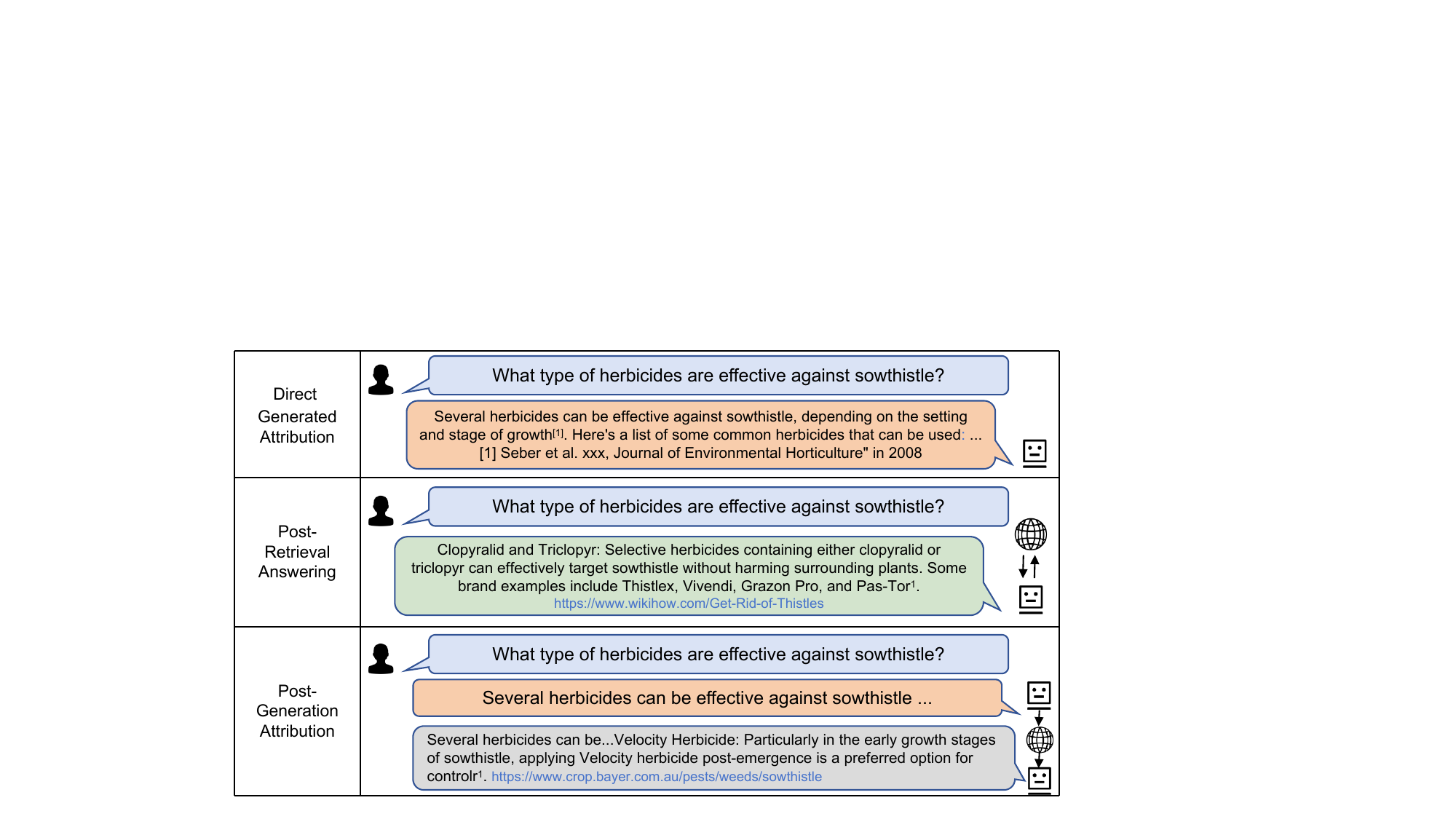}
  \caption{Three ways to attribute model-generated content. In direct model-driven attribution way, the reference document is derived from model itself and is used to cite generated answer. In post-retrieval answering way, model generates answer with citations based on the retrieved documents. In post-generation attribution way, an answer is first generated then citation and attribution are purposed.}
    \vspace{-2mm}
   \label{figexample}
\end{figure*}

The essence of the hallucination problem may stem from the fact that pre-trained models are sourced from vast, unfiltered real-world texts~\cite{Penedo2023TheRD}. These human-generated texts inherently contain inconsistencies and falsehoods. The objective of pre-training is merely to predict the next word, without explicitly modeling the veracity of the generated content. Even after utilizing reinforcement learning from human feedback~\cite{Ouyang2022TrainingLM}, models can still exhibit external hallucinations~\cite{Bai2022TrainingAH}. To address the issue of external hallucinations, researchers have begun to employ measures like external references to enhance the authenticity and reliability of chatbots~\cite{Thoppilan2022LaMDALM,Menick2022TeachingLM,nakano2021webgpt}. The distinction between explicit attribution and reinforcement learning lies not only in the need for human verification and compliance but also in recognizing that generated content might become outdated or invalid over time. As shown in Figure~\ref{figmotivation}, attribution can leverage real-time information to ensure relevance and accuracy. However, the fundamental challenge of attribution revolves around two essential requirements~\cite{Liu2023EvaluatingVI}:

\begin{enumerate}
    \item \textbf{Comprehensive attribution or citation (high recall):} Every claim and statement (except debatable or subjective text, e.g., abstained text) made by the model-generated content should be fully backed by appropriate references.
 
    \item \textbf{Sufficiency attribution or citation (high precision):} Every reference should directly support its associated claim or statement.
\end{enumerate} 
With these requirements in mind, we can break down the main ways models handle attribution into three types (see examples in Figure~\ref{figexample}):

\begin{enumerate}
    \item \textbf{Direct model-driven attribution:} The large model itself provides the attribution for its answer. However, this type often poses a challenge as not only might the answers be hallucinated, but the attributions themselves can also be~\cite{DBLP:journals/corr/abs-2305-18248}. While ChatGPT provides correct or partially correct answers about 50.6\% of the time, the suggested references were only present 14\% of the time~\cite{zuccon2023chatgpt}.

    \item \textbf{Post-retrieval answering:} This approach is rooted in the idea of explicitly retrieving information and then letting the model answer based on this retrieved data. But retrieval does not inherently equate to attribution~\cite{gao2023enabling}. Issues arise when the boundaries between internal knowledge of the model and externally retrieved information become blurred, leading to potential knowledge conflicts~\cite{Xie2023AdaptiveCO}. Retrieval can also be used as a specialized tool allowing the model to trigger it independently, similar to the \texttt{Browse with Bing} in ChatGPT~\footnote{https://openai.com/blog/chatgpt-plugins}.

    \item \textbf{Post-generation attribution:} The system first provides an answer, then conducts a search using both the question and answer for attribution. The answer is then modified if necessary and appropriately attributed. Modern search engines like Bing Chat~\footnote{https://www.bing.com/new} have already incorporated such attribution. However, studies have shown that only 51.5\% of the content generated from four generative search engines was entirely supported by their cited references~\cite{Liu2023EvaluatingVI}. This form of attribution is particularly lacking in high-risk professional fields such as medicine and law, with research revealing a significant number of incomplete attributions (35\% and 31\%, respectively); moreover, many attributions were derived from unreliable sources, with 51\% of them being assessed as unreliable by experts~\cite{Malaviya2023ExpertQAEQ}.
\end{enumerate}

Moving beyond general discussions on text hallucinations~\cite{Zhang2023SirensSI,Ye2023CognitiveMA,Rawte2023ASO}, our study delves deeper into the attribution of large language models. As shown in Figure~\ref{taxo_of_icl}, we explore its origins, the technology underpinning it, and the criteria for its assessment. Additionally, we touch upon challenges such as biases and the potential for excessive citations. We believe that by focusing on these attribution issues, we can make models more trustworthy and easier to understand. Our goal with this study is to shed light on attribution in a way that's clearer and encourages deeper thought on the topic.
\section{Task Definition}
Attribution refers to the capacity of an entity, such as a text model, to generate and provide evidence, often in the form of references or citations, that substantiates the claims or statements it produces. This evidence is derived from identifiable sources, ensuring that the claims can be logically inferred from a foundational corpus, making them comprehensible and verifiable by a general audience. Attribution itself is related to search task~\cite{Brin1998TheAO,Page1999ThePC,DBLP:conf/nips/Tay00NBM000GSCM22} where only several web pages are returned. However, the primary purposes of attribution include enabling users to validate the claims made by the model, promoting the generation of text that closely aligns with the cited sources to enhance accuracy and reduce misinformation or hallucination, and establishing a structured framework for evaluating the completeness and relevance of the supporting evidence in relation to the presented claims. The accuracy of attribution centers on \textit{whether the produced statement is entirely backed by the referenced source}.~\citet{DBLP:journals/corr/abs-2112-12870} also propose the \textit{attributed to identified sources} (AIS) evaluation framework to assess whether a particular statement is supported by provided evidence. \citet{DBLP:journals/corr/abs-2212-08037} propose attributed question answering, where the model takes a question and produces a paired response of an answer string and its supporting evidence from a specific corpus, such as paragraphs.

Formally, consider a query \(q\) (or a instruction, a prompt) and a corpus of text passages \(\mathcal{D}\). The objective of the system is to produce an output \(\mathcal{S}\), where \(\mathcal{S}\) is a set of \(n\) distinct statements: \(s_1, s_2, \ldots, s_n\). Each statement \(s_i\) is associated with a set of citations \(\mathcal{C}_i\). This set \(\mathcal{C}_i\) is defined as \(\mathcal{C}_i = \{c_{i,1}, c_{i,2}, \ldots\}\), where each \(c_{i,j}\) is a passage from the corpus \(\mathcal{D}\). For practical applications, the output from large language models can be segmented into individual statements using sentence boundaries. This approach is utilized because a single sentence typically encapsulates a coherent statement while maintaining brevity, facilitating easy verification. In terms of representation, citations may be enclosed within square brackets, for instance, \texttt{[1][2]}. It should be noted, however, that these citations can also be applied at the phrase level, rather than exclusively at the sentence level.
It is important to highlight that the task configurations discussed in this paper are distinct from the generation of citation texts found in scholarly articles or wikipedia, where the citing and cited documents are usually used as inputs~\cite{DBLP:conf/cikm/FetahuMNA16,DBLP:conf/acl/XingFW20,DBLP:journals/corr/abs-2112-01332,DBLP:journals/corr/abs-2211-07066}.

\section{Sources of Attribution}
\label{sec:Sources_of_Attribution}
\subsection{Pre-training Data}
\label{sec:Pre-training_Data}
LLMs are typically trained on extensive corpora collected from various sources, predominantly the web. This vast amount of pre-training data forms the bedrock on which these models develop their understanding and capabilities. However, due to the scale of the data involved, manual inspection is often unfeasible, leading to potential inaccuracies, biases, and other undesirable artifacts in the data~\cite{DBLP:conf/acl/PiktusAVLDLJR23}. Despite these challenges, LLMs tend to perform well on a wide array of downstream tasks, even with little to no task-specific tuning. This performance hints at the ability of models to either memorize or reason through patterns present in the data. However, the specific patterns or the extent to which they are memorized or reasoned through, especially in different downstream tasks, remain somewhat elusive.

The concept of attribution in this context refers to tracing back the behavior of the model on a particular task to specific portions of the pre-training data~\cite{DBLP:journals/corr/abs-2205-12600, weller2023according}. By identifying a subset of pre-training data that significantly influences the model behavior on a downstream task, researchers aim to provide a clearer understanding of how the pre-training data impacts the model's performance~\cite{DBLP:conf/acl/HanSMTCW23}. This kind of attribution is essential for interpreting the model, providing insights into whether the model is capturing task-relevant patterns or merely memorizing data. Furthermore, it aids in enhancing the trustworthiness of the model by offering a clearer picture of how the model operates and what sources of data significantly contribute to its performance. Through such attribution methodologies, researchers aim to bridge the understanding gap, offering a pathway towards better interpretability, trustworthiness, and eventually, the improvement of LLMs in handling various NLP tasks.

\subsection{Out-of-model Knowledge}
\label{sec:Out-of-model_Knowledge}
This source reveals methods to leverage out-of-model knowledge (e.g., web, knowledge graph) for attribution to enhance  capabilities of models~\cite{DBLP:conf/emnlp/0001PCKW21,li2023verifiable}. Primary among these methods is the retrieval-augmented generation technique~\cite{DBLP:conf/nips/LewisPPPKGKLYR020} which uses an encoder-decoder mechanism to encode questions and decode answers, augmented with documents or passages from extensive unstructured datasets. Furthermore, retrieval-enhanced language models are highlighted, which improve performance by fetching $k$-most similar training contexts or generating search queries to obtain relevant documents from external sources~\cite{DBLP:conf/icml/BorgeaudMHCRM0L22}. These methodologies, along with a mentioned post-processing method to utilize retrieved knowledge without additional training or fine-tuning, represent critical pathways for attributing LLM responses or generated text to external knowledge, aiming to make the outputs of LLMs to identifiable and verifiable external knowledge sources~\cite{DBLP:conf/eacl/IzacardG21,DBLP:journals/corr/abs-2301-00303}.

\section{Datasets for Attribution}
\label{sec:Datasets_for_Attribution}
As an information-seeking task, datasets for attribution are often built in the form of question answering or text summarization (see Table~\ref{tabledatasets}). Several benchmarks are proposed based on existing QA datasets by proposing methods to evaluate the performance of attribution, as the golden citation annotation is not a necessity.
\citet{nakano2021webgpt} built a long-form QA dataset with web search results. After that \citet{qin-etal-2023-webcpm} built a similar Chinese dataest for the same purpose. However, these datasets are not directly built for verifying citations, but for factual accuracy.
More recently, several works \cite{DBLP:journals/corr/abs-2304-04358,gao2023enabling,hagrid,Malaviya2023ExpertQAEQ,li2023verifiable} focus on measuring and improving the accuracy of citations in generated text based on a given set of quotes, varying on question domain and citation granularity.


\noindent\textbf{Question Domain.} Most relevant datasets are designed for open-domain. However, ExpertQA~\cite{Malaviya2023ExpertQAEQ} chose 32 domain-specific scenarios, some of them are high-stakes fields, and brought domain experts in the loop. BioKaLMA~\cite{li2023verifiable} focused on biography domain for its practical application and convenient evaluation.

\noindent\textbf{Attribution Granularity.} There are two kinds of citation granularity in recent works: entity and sentence. The entity level attribution are more fine-grained, sentence level attribution requires citation for every completed sentence. Among them,  ExpertQA~\cite{Malaviya2023ExpertQAEQ} and BioKaLMA~\cite{li2023verifiable} make attribution at entity level, whereas other method make attribution at sentence level.

\begin{table*}
  \centering
  \scalebox{0.8}{
  \begin{tabular}{cccccc}
    \toprule
    Dataset & Task & Domain & Source & Structure & Granularity \\
    \hline
    WebGPT~\cite{nakano2021webgpt} & QA & Open-domain & Web Pages & Unstructured & Sentence \\
    WebCPM~\cite{qin-etal-2023-webcpm} & QA & Open-domain & Web Pages & Unstructured & Sentence \\
    CiteBench~\cite{https://doi.org/10.48550/arxiv.2212.09577} & Summ. & Scientific & Scientific Paper & Unstructured & Sentence \\
    HAGRID~\cite{hagrid} & QA & Open-domain & Wikipedia & Unstructured & Sentence \\
    ALCE~\cite{gao2023enabling} & QA & Open-domain & Wikipedia+Sphere & Unstructured & Sentence \\
    BioKaLMA~\cite{li2023verifiable} & QA & Biography & Wikipedia & Structured & Entity \\
    ExpertQA~\cite{Malaviya2023ExpertQAEQ} & QA & Specific domains & Wikipedia & Unstructured & Entity \\
    \bottomrule
  \end{tabular}
  }
  \caption{Comparsion between different datasets for attribution.}
  \vspace{-2mm}
  \label{tabledatasets}
\end{table*}


\section{Approaches to Attribution}
\label{sec:Approaches_to_Attribution}

\subsection{Direct Generated Attribution}
\label{sec:Direct_Generated_Attribution}


Attribution from parametric knowledge can help reduce hallucination and improve the truthfulness of generated text. By asking models to do self-detection and self-attribution, some researches found that the generated texts are more grounded on facts and additionally improve performance on down-stream tasks~\cite{DBLP:conf/iclr/Sun0TYZ23}.

Recently, researchers found that large language models can not provide knowledge sources or evidences clearly when answering domain-specific knowlege-based questions~\cite{DBLP:conf/acl/PeskoffS23,zuccon2023chatgpt,GRAVEL2023226}. In most cases, models can only provide a knowledge source that is loosely related to the keywords in questions or irrelevant to current topics. Even if the model answered the question correctly, the evidence it provided is still likely to have mistakes.

\citet{weller2023according} tries to ground model's generated text to its pre-training data by proposing according-to prompting, who finds the method can affect model's groundedness and therefore affect performance on information-seeking tasks.

\citet{anonymous2023learning} introduces an intermediate planning module, asking the model to generate a series of questions as blueprints to the current question. The model first propose a blueprint and then combines the texts which are generated based on the blueprint questions as the final answer. The blueprint models allow for different forms of attribution during each question answering step, which can expect to be more explainable. 







\subsection{Post-retrieval Answering}
\label{sec:Post-retrieval_Answering}

Numerous studies have delved into the post-retrieval answering strategy for attribution~\cite{danqichen17,KentonLee19,colbert}. \citet{DBLP:journals/corr/abs-2303-14337} introduce the SmartBook framework, which aims to generate structured situation reports incorporating factual evidence through rich links. The framework autonomously identifies crucial questions for situation analysis and extracts pertinent information to compose the report. Each question is addressed with concise summaries containing tactical details from pertinent claims, supported by reliable and trustworthy factual evidence.
To tackle the issue of misalignment between user queries and stored knowledge, where LLMs struggle to correlate questions with the appropriate grounding, MixAlign \citep{DBLP:journals/corr/abs-2305-13669} presents a framework that combines automatic question-knowledge alignment with user clarifications. This approach effectively mitigates language model hallucination.
To assess the adequacy of document support for an answer, LLatrieval~\cite{li2023llatrieval} updates the retrieval results until it confirms that the retrieved documents can sufficiently support the answer to the question. This iterative verification process significantly enhances attribution accuracy by ensuring that the generated response is backed by verifiable evidence.
Similarly, Self-RAG~\cite{DBLP:journals/corr/abs-2310-11511} trains an arbitrary language model to generate reflection-specific tokens after knowledge retrieval, thereby augmenting the attribution of retrieved passages.
Furthermore, the Search-in-the-chain (SearChain) \citep{DBLP:journals/corr/abs-2304-14732} introduces a novel approach to address the challenges posed by incorrect knowledge retrieved by information retrieval systems, which can mislead LLMs or disrupt their reasoning chains. It verifies and corrects answers within the global reasoning chain, known as Chain-of-Query (CoQ), while also identifying missing knowledge in CoQ. These operations significantly enhance the attribution accuracy of LLMs in complex knowledge-intensive tasks, improving their reasoning ability and knowledge utilization.

\subsection{Post-Generation Attribution}
\label{sec:Post-Generation_Attribution}

\begin{figure}[t!]
\centering
\includegraphics[width=0.9\linewidth]{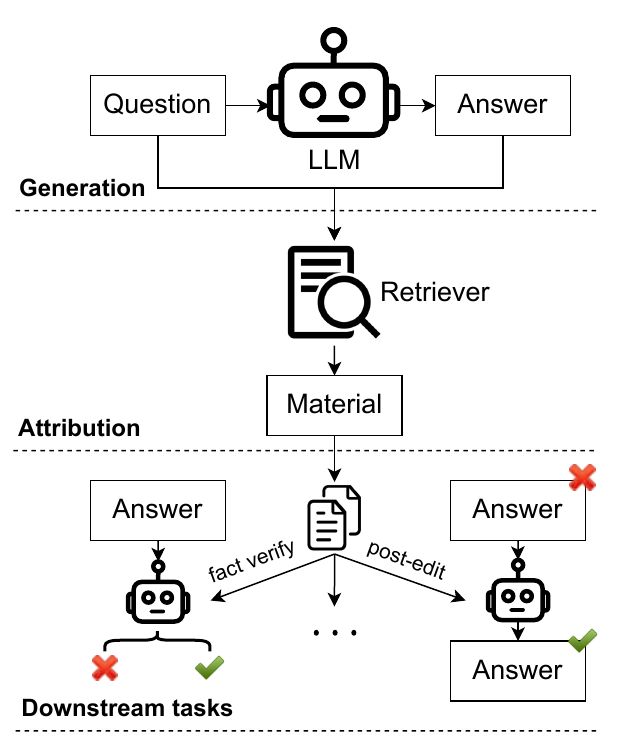}
  \caption{Workflow of post-generation attribution. Retrieval is performed after an answer being generated. The retrieved documents are used to perform citation and attribution, subsequently used to do fact verification and post-editing.}
    \vspace{-4mm}
   \label{6.3-workflow}
\end{figure}

In order to facilitate accurate attribution without compromising the robust benefits offered by recent-generation models, some researches aim at attribution after generation, which employ search engines or document retrieval systems for searching evidence base on the input questions and generated answers. This approach allows researchers to assess or improve the factuality of answers without needing to access the model's parameters directly. The workflow of post-generation attribution is illustrated in Figure~\ref{6.3-workflow}.
RARR~\citep{DBLP:conf/acl/GaoDPCCFZLLJG23} autonomously identifies the attribution for the output of any text generation model, and performs post-editing to rectify unsupported content, whilst striving to retain the original output to the greatest extent feasible.
In the work of \citet{DBLP:journals/corr/abs-2309-11392}, materials are retrieved from the corpus base on coarse-grained sentences or fine-grained factual statements. These retrieved materials are then utilized to prompt the LLM to verify the consistency between the generated responses and the retrieved material, and to make necessary edits to reduce the hallucinations.
\citet{DBLP:journals/corr/abs-2305-11859} introduces a fully-automated pipeline designed to verify complex political claims, which is achieved by retrieving raw evidence from the web, generating claim-focused summaries and utilizing them for claim verification.



\begin{table*}
  \centering
  \scalebox{0.8}{      
  \begin{tabular}{cccccc}
    \toprule
    System & Model Training & Evidence Type & Citation Type & Attribution Integration \\
    \hline
    LaMDA~\cite{Thoppilan2022LaMDALM} & Multi-task FT & Snippets & URLs & Appended \\
    WebGPT~\cite{nakano2021webgpt} & SFT + RL & Well-curated documents & Documents & Embedded \\
    GopherCite~\cite{Menick2022TeachingLM} & SFT + RL & Long documents & Documents & Embedded\\
    Sparrow~\cite{DBLP:journals/corr/abs-2209-14375} & RL & Well-curated documents & Documents & Appended\\
    \bottomrule
  \end{tabular}
  }
  \caption{Features of different attribution systems.}
  \vspace{-2mm}
  \label{tablesystem}
\end{table*}

\subsection{Attribution Systems}
\label{sec:Attribution_Systems}

\citet{Thoppilan2022LaMDALM} introduced LaMDA, a dialogue-focused language model. While enlarging the model improves its quality, it does not necessarily enhance safety and accuracy. By fine-tuning LaMDA with annotated data and enabling it to access external knowledge, they significantly improved its safety and factual grounding. The grounding challenge of this study aims to generate responses based on credible external sources instead of merely plausible ones.
The WebGPT model~\cite{nakano2021webgpt} based on GPT-3 is trained to search and navigate the web and is fine-tuned for answering long-form questions in a web-browsing environment. For human evaluation of its factual accuracy, the model is required to gather references while browsing Microsoft Bing to support its answers. This ensures that the answers provided have a basis or attribution from credible web sources.
Similarly, GopherCite~\cite{Menick2022TeachingLM} trained with reinforcement learning references evidence from multiple documents or a single user-provided document and refrains from answering when uncertain. Human evaluations show that GopherCite produces high-quality responses 80\% at most. Nonetheless, citation alone is not a complete solution for ensuring safety and trustworthiness, as evidence-backed claims can still be false.
Sparrow~\cite{DBLP:journals/corr/abs-2209-14375} is trained to search the internet using Google Search to provide more accurate answers, allowing it to reference the latest information. In the user interface, evidence used by the model is displayed alongside its response, offering raters a means to validate the correctness of answer. To train the model in searching and using evidence, a preference model is used based on human judgments. Through human evaluation, it was found that responses with evidence were deemed plausible and supported 78\% of the time. 
Comparisons between different systems are shown in Table~\ref{tablesystem}.

\section{Attribution Evaluation}
\label{sec:Attribution_Evaluation}

\begin{table*}
  \centering
  \scalebox{0.8}{      
  \begin{tabular}{ccc}
    \toprule
    Evaluation Metrics & Evaluation Method & Description \\
    \hline
    Recall, Precision & Automatic, Statistics, Model-based & \makecell[c]{binary categorization based on NLI models}    \\
    EM, BLEU, ROUGE & Automatic, Statistics & metrics for downstream tasks \\
    QUIP-Score~\cite{weller2023according} & Automatic, Statistics & \makecell[c]{character-level n-gram metrics}\\
    ~\citet{Liu2023EvaluatingVI} & Human & fluency, perceived utility \\
    AttrScore~\cite{DBLP:journals/corr/abs-2305-06311} & Human & attributability, extrapolatory, contradiction \\
    \bottomrule
  \end{tabular}
  }
  
  \caption{Comparison between different evaluation metrics for attribution.}
  \label{tablemetrics}
\end{table*}

\noindent\textbf{Human Evaluation.}\label{sec:Human Evaluation}
To detect attribution errors, current attributed LLMs predominantly depend on human evaluation, a process that is both costly and time-intensive~\cite{nakano2021webgpt, kazemi-etal-2023-lambada, chen2023understanding}. For example, the typical cost of annotating a single (query, answer, reference) example stands at around \$1~\cite{Liu2023EvaluatingVI}. In practical applications of attributed LLMs, the responsibility falls on users to be cautious of attributions and to undertake manual verification, imposing a significant responsibility on them.

\noindent\textbf{Categorization-Based Evaluation.}
For the sake of clarity, earlier research mainly employed binary categorization by repurposing other NLP tasks (e.g., natural language inference) to determine whether an answer is supported by a reference or not (attributable or not)~\cite{DBLP:journals/corr/abs-2112-12870,DBLP:journals/corr/abs-2212-08037,gao2023enabling, DBLP:journals/corr/abs-2305-14332}. ~\citet{Liu2023EvaluatingVI} carried out a human assessment to evaluate the veracity of responses from generative search engines, categorizing the degree of reference support into full, partial, or no support. Building on this, ~\citet{DBLP:journals/corr/abs-2305-06311} introduced a refined categorization of attribution: 1) attributable--where the reference entirely backs the generated statement; 2) extrapolatory--where the reference offers insufficient backing for the statement; and 3) contradictory--where the statement directly opposes the referenced citation.

\noindent\textbf{Quantitative Evaluation Metrics.}
Assessment of attribution quality is approached from three distinct angles~\cite{li2023verifiable}: 1) Correctness--evaluating the alignment of generated text with the provided sources; 2) Precision--measuring the percentage of generated attributions pertinent to the question at hand; and 3) Recall--assessing the scope to which generated attributions capture crucial knowledge. Moreover, the F1-Score is derived from the Precision and Recall metrics.
\citet{Thoppilan2022LaMDALM} introduces citation accuracy as the frequency with which the model refers to web sources for its assertions, excluding widely recognized truths. The QUIP-Score~\cite{weller2023according}, an n-gram overlap metric, is founded on swift membership inquiries and evaluates the extent to which a section is comprised of exact spans within a text corpus.

In summary, the detection of attribution errors is paramount in current LLMs. While human evaluations provide in-depth insights, their costly and time-consuming nature emphasizes the growing appeal for automated methods. Future research is expected to refine these methods, ensuring their practicality and reliability in real-world applications.

\section{Discussion}
\subsection{Attribution Error Analysis}
Attribution error has several forms.  In this study, we systematically categorize these errors into three primary types, as outlined in Table~\ref{tableerrors}, while acknowledging the possibility of other error types.

\noindent\textbf{Granularity Error.} For ambiguous questions, the answer may involve multiple aspects. In this case, the retrieved multi-document may contain complex and diverse information. Thus the answer is complex and hybrid, leading to insufficient citation.

\noindent\textbf{Mistaken synthesis.} Models may mix up relationships between entities and events when several complex documents are provided. The citation should be faithful to the generated text and cite all the references.

\noindent\textbf{Hallucinated Generation.} The reference documents may be irrelevant or not relevant to the question, or the model has conflicts between external documents and parameter knowledge. The answer will be hallucinated and the citation is inaccurate.

\subsection{Limitations of Attributions}
\label{sec:limitation}
Attribution in LLMs is fraught with inherent difficulties. One primary challenge is discerning when and how to attribute. Differentiating between general knowledge, which may not require citations, and specialized knowledge, which should ideally be attributed, is a nuanced task. This gray area can lead to inconsistencies in attribution~\cite{DBLP:journals/corr/abs-2307-02185}. And LLMs now do not have ability to attribute parameter knowledge of itself~\cite{litschko2023establishing}.
Another limitation is the potential inaccuracy in attributions~\cite{Liu2023EvaluatingVI}. LLMs might link content to irrelevant or incorrect sources. This misattribution can confuse users, leading them wrong and affecting the reliability of the information presented. For example, an LLM in the medical field could wrongly associate faulty medical guidance with a trustworthy reference, which might guide users towards detrimental health choices.
Furthermore, the fluidity of knowledge means that while some information remains static, other data evolves and changes over time~\cite{DBLP:journals/corr/abs-2305-14251}. Consequently, some attributions made by LLMs may quickly become outdated, especially in rapidly advancing domains, such as computer science and clinical medicine.  Additionally, we recommend readers refer to \S4.1 in~\citet{Menick2022TeachingLM}.

\subsection{Challenges for Attributions}
Despite the potential solutions on the horizon, implementing these improvements is laden with challenges. One such challenge is excessive attribution or over attribution~\cite{DBLP:journals/corr/abs-2307-02185,Liu2023EvaluatingVI}. 
If LLMs give credit too often, users might get overwhelmed with too much information, confusing them and making it difficult to tell what is important and relevant from what is not.
At the same time, there is a real chance of LLMs accidentally revealing private information. Finding a balance between clear attribution and protecting private details is a tricky task.
Bias is another big challenge. LLMs might unintentionally lean towards some sources or kinds of information, pushing certain views while ignoring others. To tackle this bias, we need to use varied training data and improve the methods used for giving credit~\cite{DBLP:journals/corr/abs-2306-11644}.

Lastly, the shadow of incorrect information is ever-present. Without solid validation measures, LLMs could potentially spread wrong or misleading details, undermining the reliability of the information landscape. 
Future models should recognize ambiguous references and refrain from making statements when the evidence is not clear, instead of presenting unfounded claims.
Overall, though LLMs seem to be on a positive path, they face many obstacles and doubts. Proper credit is not just a side aspect; it is vital to the growth, approval, and effectiveness of LLMs. Guaranteeing correct and reliable credits, while promoting new ideas, will definitely influence the future of LLMs.
\subsection{Future Directions for Attributions}
\noindent\textbf{Continuous Refreshment of LLMs.}
A promising direction for upcoming advancements is to create a system that consistently refreshes the information of LLMs~\cite{Thoppilan2022LaMDALM,nakano2021webgpt}, akin to how search engines update their databases. This approach not only ensures up-to-date content for attribution but also offers a platform for continuous learning and adaptation.

\noindent\textbf{Enhancing the Reliability of LLM Outputs.}
Another pivotal direction entails boosting the trustworthiness of LLM outputs. This can be achieved by incorporating rigorous systems that evaluate the credibility and accuracy of the sources to which they attribute information~\cite{DBLP:journals/corr/abs-2303-08896,DBLP:journals/corr/abs-2305-14251}. Ensuring reliable and consistent sources will instill greater confidence in users about the content generated. As the adoption of LLMs expands across various domains, the reliability of their outputs becomes critical for informed decision-making in diverse sectors.

\noindent\textbf{Balancing Creativity with Proper Credit Attribution.}
Furthermore, LLMs are recognized for their creative content generation. Striking a balance between this inventive ability and proper credit-giving is a delicate act that needs investigation. While creativity is one of the significant strengths of LLMs, it is vital to ensure that the generated content remains trustworthy and rooted in factual bases. The aim is to make sure LLMs acknowledge sources without hindering their creative potential. Balancing these two aspects can foster an environment where users both benefit from the model and trust its outputs.

\section*{Limitation}
While language models have the capability to cite their sources, undeniably enhance their utility, several limitations arise that need careful consideration (cf. Section~\ref{sec:limitation}). 
Our paper, in its current form, does not provide a solution to navigate such complex territory. It is important to address these limitations in future works and to continually educate users about the potential pitfalls of relying solely on machine-generated text.
\normalem
\bibliography{acl_latex}
\appendix
\clearpage
\section{Attribution Before the Era of Large Language Model}
\subsection{Related Natural Language Processing Tasks}

The relationship between attribution tasks and other Natural Language Processing (NLP) tasks manifests in the overarching goal of understanding, evaluating, and leveraging the content retrieved or generated in response to particular stimuli such as questions or claims. Here is an exploration of how attribution tasks are intertwined with other NLP tasks, anchored on the retrieval of related content:

\noindent\textbf{Open-domain question answering:} Both tasks hinge on retrieving pertinent documents or information to address a posed question or claim. While open-domain QA zeroes in on the accuracy and relevance of the answer, attribution tasks scrutinize whether the answer or generated text can be accurately traced back to the retrieved documents \cite{danqichen17, DBLP:journals/corr/abs-2212-08037}.

\noindent\textbf{Fact-checking \& Claim verification (subtask of fact checking):} Fact-checking and attribution tasks both necessitate the retrieval of external evidence to validate a claim or generated text.
The emphasis in fact-checking is on verifying the truthfulness of a claim, whereas attribution tasks focus on the correct attribution of generated text to the sourced evidence \cite{DBLP:conf/naacl/ThorneVCM18}. 
On the other hand, attribution tasks and claim verification both center around validating information against reference or sourced material, yet they serve different purposes. Attribution ensures that generated text or answers accurately reflect the provided references, while claim verification assesses the truthfulness of a claim based on evidence or source material~\cite{DBLP:journals/tacl/GuoSV22}. Both tasks necessitate the retrieval of related content for verification~\cite{wang2023explainable}, making them inherently reliant on the accuracy and relevance of the retrieved material. claim verification pivotal in fact-checking and misinformation detection, they share the fundamental objective of endorsing the accuracy and trustworthiness of information by juxtaposing it against a reference.

\noindent\textbf{Natural Language Inference (NLI):} Both tasks engage in evaluating the relationship between two snippets of text; however, NLI concentrates on logical entailment, contradiction, or neutrality, while attribution evaluates the substantiation provided by references for generated text \cite{DBLP:conf/emnlp/BowmanAPM15}.

\noindent\textbf{Summarization:} Summarization and attribution tasks both generate condensed or altered text and necessitate a check on the fidelity of the generated text to the original or sourced content.
Attribution in summarization is pivotal to averting hallucinations (generation of false or unsupported information) and ensuring the summary accurately mirrors the input text \cite{DBLP:journals/csur/JiLFYSXIBMF23}.

The commonality among these tasks lies in the requisite to retrieve, analyze, and validate content against some form of reference material, be it external evidence, retrieved documents, or a different segment of text. The capacity to retrieve related content forms a cornerstone for these tasks, enabling the necessary comparisons and evaluations to ascertain accuracy, relevance, and correct attribution.
\subsection{Interpretability of NLP Models}

Interpretability (e.g., feature attribution) dives into understanding which parts of the input (e.g., words or phrases) are crucial for a model's decision or output~\cite{DBLP:conf/acl/LiuA19,DBLP:conf/aaai/LiHCXTZ22}. It helps in identifying the importance of different features in the input data concerning the model's performance. Compared to feature attribution, explicit attribution for LLMs serves as a conduit to trace the sources of the information they generate, which is pivotal for accountability, especially in critical domains like healthcare or finance. It enables verifiability, allowing users or other systems to check the accuracy and reliability of the information provided. Trustworthiness is also fostered through explicit attribution, as users are more likely to trust the model if they know where the information is coming from. Additionally, it plays a role in interpretability, aiding users in understanding how the model arrives at certain conclusions by revealing the sources of information.` This alignment with interpretability objectives helps in making the model's decision-making process more transparent and comprehensible.

Attribution and interpretability, though interconnected, serve distinct purposes. Attribution specifically refers to the process of tracing back the generated information or decisions of a model to its source material or input features, providing a clear reference or basis for the output. On the other hand, interpretability is a broader concept encompassing the understanding of how a model processes input data to arrive at a particular output~\cite{DBLP:journals/corr/abs-2202-01875}, making the inner workings of the model transparent and comprehensible to users. While attribution can be seen as a component or a specific form of interpretability, aiding in understanding and trusting the model's outputs by providing source references, interpretability dives deeper into elucidating the model's behavior, decision-making process, and the significance of different input features in those decisions, thus fostering a comprehensive understanding and trust in the model. Note that the evaluation of attribution focuses on the verifiability of the answer based on the reference, similar to \textit{faithfulness} evaluation of explanations~\cite{jacovi-goldberg-2020-towards}. It does not measure the relevance about whether the answer correctly responds to the query, similar to \textit{plausibility} in~\citet{jacovi-goldberg-2020-towards}.

\begin{table*}
  \centering
  \small
  \resizebox{\linewidth}{!}{   
  \begin{tabular}{cp{12cm}}
    \toprule
    \multirow{4}{*}{Granularity Error} & \textbf{Question: }Where was the biggest gold nugget ever found?\\ 
                                            & \textbf{References: } [doc1] The largest gold nugget ever {\color{blue}found in California} was found in August 1869 in Sierra Buttes by five partners. [doc2] Considered by most authorities to be the biggest gold nugget ever found, the Welcome Stranger was found at Moliagul, Victoria, Australia in 1869. [doc3] Pepita Canaã, was found on September 13, 1983 by miners at the Serra Pelada Mine {\color{blue}in the State of Para, Brazil}. \\ 
                                            & \textbf{Answer: }Most authorities consider the Welcome Stranger to be the biggest gold nugget ever found, found in 1869 in Moliagul, Victoria, Australia[2]. {\color{blue}The biggest gold nugget found in Brazil was found in the State of Para and in United States was found in Sierra Buttes }{\color{red}[3].}\\ 
                                            & \textbf{Explanation: } The second sentence in generated answer has two facets( the biggest nugget in Barzil, and the biggest nugget in United States). However, only document 3 is cited. \\
                                            
                                            \hline
    \multirow{4}{*}{Mistaken Synthesis} & \textbf{Question: }What's the difference in the portrayal of Ares in Roman mythology and Greek mythology? \\ 
                                            & \textbf{References: }[doc1] {\color{blue}In Roman mythology}, he is called Mars. The Latin name for Mars derives from his Roman name; "Tuesday" comes from the corresponding deity Tyr in Norse mythology...[doc2] Ares is the god of war in ancient Greek mythology, {\color{blue}one of the twelve Olympian gods}. He is the son of Zeus and Hera, or alternatively, he was born after Juno (Hera's Roman name) swallowed a violent-eyed giant snake...\\ 
                                            & \textbf{Answer: }In Roman mythology, Ares is called Mars, {\color{red}one of the twelve main gods of Rome}, representing the symbols of power and war{\color{red}[1]}. In Greek mythology, Ares is also one of the twelve Roman gods, representing the symbols of power and war[2]. \\ 
                                            & \textbf{Explanation: }The answer mistakenly mixes up Olympian gods and Roman mythology thus both document 1 and document 2 should be treat as referenced documents. \\
                                            \hline
    \multirow{4}{*}{Hallucinated Generation} & \textbf{Question: }How many times have the lakers won the finals?\\ 
                                          & \textbf{References: }[doc1] ... won six times in the 1960s, three times in the 1980s, in 2008, and in 2010... [doc2] ...and Lakers met in the Finals for the first time since 1987, with the Celtics winning the series 4-2. They faced off once again in the 2010 NBA Finals, which the Lakers won in 7 games...\\ 
                                          & \textbf{Answer: }The Los Angeles Lakers have won the NBA Finals {\color{red}five times, in 1980, 1982, 1985, 1987, and 2010[1]}.\\
                                          & \textbf{Explanation: }The answer is incorrect as the generation is not grounded on reference documents. Hallucination may happen because of the conflict between reference documents and parameter knowledge, thus the citation will result in failure.\\
    \bottomrule
  \end{tabular}
  }
  
  \caption{List of different attribution errors types and example instance. }
  \vspace{-2mm}
  \label{tableerrors}
\end{table*}

\end{document}